\documentclass[runningheads]{llncs}

\usepackage{graphicx}
\usepackage{amsfonts}
\usepackage{algorithm}
\usepackage{url}
\usepackage[noend]{algpseudocode}
\usepackage[table,xcdraw]{xcolor}
\usepackage{lipsum}
\usepackage{bbding}

\begin{document}

\title{Interpretable Option Discovery using Deep Q-Learning and Variational Autoencoders}
\titlerunning{Deep Variational Q-Networks for Options Discovery}

\author{Per-Arne Andersen\textsuperscript{(\Envelope)} \and Morten Goodwin \and Ole-Christoffer Granmo}
%
\authorrunning{P.-A. Andersen et al.}
\institute{Department of ICT, University of Agder, Grimstad, Norway\\
		\email{\{per.andersen,morten.goodwin,ole.granmo\}@uia.no}}

\maketitle              

\begin{abstract}
Deep Reinforcement Learning (RL) is unquestionably a robust framework to train autonomous agents in a wide variety of disciplines. However, traditional deep and shallow model-free RL algorithms suffer from low sample efficiency and inadequate generalization for sparse state spaces. The options framework with temporal abstractions \cite{Sutton1999} is perhaps the most promising method to solve these problems, but it still has noticeable shortcomings. It only guarantees local convergence, and it is challenging to automate initiation and termination conditions, which in practice are commonly hand-crafted.

Our proposal, the Deep Variational Q-Network (DVQN), combines deep generative- and reinforcement learning. The algorithm finds good policies from a Gaussian distributed latent-space, which is especially useful for defining options. The DVQN algorithm uses MSE with KL-divergence as regularization, combined with traditional Q-Learning updates. The algorithm learns a latent-space that represents good policies with state clusters for options. We show that the DVQN algorithm is a promising approach for identifying initiation and termination conditions for option-based reinforcement learning. Experiments show that the DVQN algorithm, with automatic initiation and termination, has comparable performance to Rainbow and can maintain stability when trained for extended periods after convergence.
\keywords{Deep Reinforcement Learning \and Clustering \and Options \and Hierarchical Reinforcement Learning \and Latent-space representation}
\end{abstract}

\section{Introduction}
The interest in deep Reinforcement Learning (RL) is rapidly growing due to significant progress in several RL problems \cite{Arulkumaran2017}. Deep RL has shown excellent abilities in a wide variety of domains, such as video games, robotics and, natural language progressing \cite{Silver2017,Mnih2015,Levine2015}. Current trends in applied RL has been to treat neural networks as black-boxes without regard for the latent-space structure. While unorganized latent-vectors are acceptable for model-free RL, it is disadvantageous for schemes such as options-based RL. In an option-based RL, the policy splits into sub-policies that perform individual behaviors based on the current state of the agent. A sub-policy, or option, is selected with initialization criteria and ended with a termination signal. The current state-of-the-art in option-based RL primarily uses hand-crafted options. Option-based RL algorithms work well for simple environments but have poor performance in more complicated tasks. There is, to the best of our knowledge, no literature that addresses good option selection for difficult control tasks. There are efforts for making automatic options selection \cite{Stolle2004}, but no method achieves notable performance across various environments.

This paper proposes a novel deep learning architecture for Q-learning using variational autoencoders that learn to organize similar states in a vast latent-space. The algorithm derives good policies from a latent-space that feature interpretability and the ability to classify sub-spaces for automatic option generation. Furthermore, we can produce human-interpretable visual representations from latent-space that directly reflects the state-space structure. We call this architecture DVQN for deep Variational Q-Networks and study the learned latent-space on classic RL problems from the Open AI gym \cite{Brockman2016a}.

The paper is organized as follows. Section \ref{sec:bg} introduces preliminary literature for the proposed algorithm. Section \ref{sec:dvqn} presents the proposed algorithm architecture. Section \ref{sec:results} outlines the experiment setup and presents empirical evidence of the algorithm performance. Section \ref{sec:related_work} briefly surveys work that is similar to our contribution. Finally, Section \ref{sec:conclusion} summarises the work of this paper and outlines a roadmap for future work.

\section{Related Work}
\label{sec:related_work}
There are numerous attempts in the literature to improve interpretability with deep learning algorithms, but primarily in the supervised cases. \cite{Zhang2018} provides an in-depth survey of interpretability with Convolutional Neural Networks (CNNs). Our approach is similar to the work of \cite{Wang2019a}, where the authors propose an architecture for visual perception of the DQN algorithm. The difference, however, is primarily our focus on the interpretability of the latent-space distribution via methods commonly found in variational autoencoders. There are similar efforts to combine Q-Learning with Variational Autoencoders, such as \cite{Tang2017,Huang2020}, and shows promising results theoretically but with limited focus on interpretability. \cite{Annasamy2018} did notable work on interpretability among using a distance KL-distance for optimization but did not find convincing evidence for a deeper understanding of the model. The focus of our contribution deviates here and finds significant value in a shallow and organized latent-space. 

\textbf{Options}
The learned latent-space is valuable for the selection of options in hierarchical reinforcement learning (HRL). There is increasing engagement in HRL research because of several appealing benefits such as sample efficiency and model simplicity \cite{Barto2003}. Despite its growing attention, there are few advancements within this field compared to model-free RL. The options framework \cite{Sutton1999} is perhaps the most promising approach for HRL in terms of intuitive and convergence guarantees. Specifically, the options framework defines semi-Markov decision processes (SMDP), which is an extension of the traditional MDP framework \cite{Younes2004}. SMDP features temporal abstractions where multiple discrete time steps are generalized to a single step. These abstract steps are what defines an option, where the option is a subset of the state-space. In the proposed algorithm, the structure of the latent-space forms such temporal abstractions for options to form.

\section{Background}
\label{sec:bg}

The algorithm is formalized under conventional Markov decision processes tuples \(<S, A, P, R, \gamma>\) where \(S\) is a (finite) set of all possible states, \(A\) is a (finite) set of all possible actions, \(P\) defines the probabilistic transition function \(P(S_{t+1} = s’ | s, a)\) where \(s\) is the previous state,  and \(s’\) is the transition state. \(R\) is the reward function \(R(r_{t+1} | s, a)\). Finally, the \(\gamma\) is a discount factor between \(\gamma \in [ 0 \dots 1 ]\) that determines the importance of future states. Lower \(\gamma\) values decrease future state importance while higher values increase.

\section{Deep Variational Q-Networks}
\label{sec:dvqn}
Our contribution is a deep Q-learning algorithm that finds good policies in an organized latent space from variational autoencoders. \footnote{The code will be published upon publication.} Empirically, the algorithm shows comparable performance to traditional model-free deep Q-Networks variants. We name our method the \textbf{D}eep \textbf{V}ariational \textbf{Q}-\textbf{N}etwork (DVQN) that combines two emerging algorithms, the variational autoencoder (VAE) \cite{Kingma2013} and deep Q-Networks (DQN) \cite{Mnih2015}.

In traditional deep Q-Networks, the (latent-space) hidden layers are treated as a black-box. On the contrary, the objective of the variational autoencoder is to reconstruct the input and \textbf{organize} the latent-vector so that similar (data) states are adjacently modeled as a Gaussian distribution.

In DQN, the latent-space is sparse and is hard to interpret for humans and even option-based machines. By introducing a VAE mechanism into the algorithm, we expect far better interpretability for creating options in RL, which is the primary motivation for this contribution. Variational autoencoders are, in contrast to deep RL, involved with the organization of the latent-space representation, and commonly used to generate clusters of similar data with t-SNE or PCA \cite{Zheng2016}. The DVQN algorithm introduces three significant properties. First, the algorithm fits the data as a Gaussian distribution. This reduces the policy-space, which in practice reduces the probability of the policy drifting away from global minima. Second, the algorithm is generative and does not require exploration schemes such as \(\epsilon\)-greedy because it is done in re-parametrization during training. Third, the algorithm can learn the transition function and, if desirable, generate training data directly from the latent-space parameters, similar to the work of \cite{Ha2018a}. 

\begin{figure*}
    \centering
    \includegraphics[width=\linewidth]{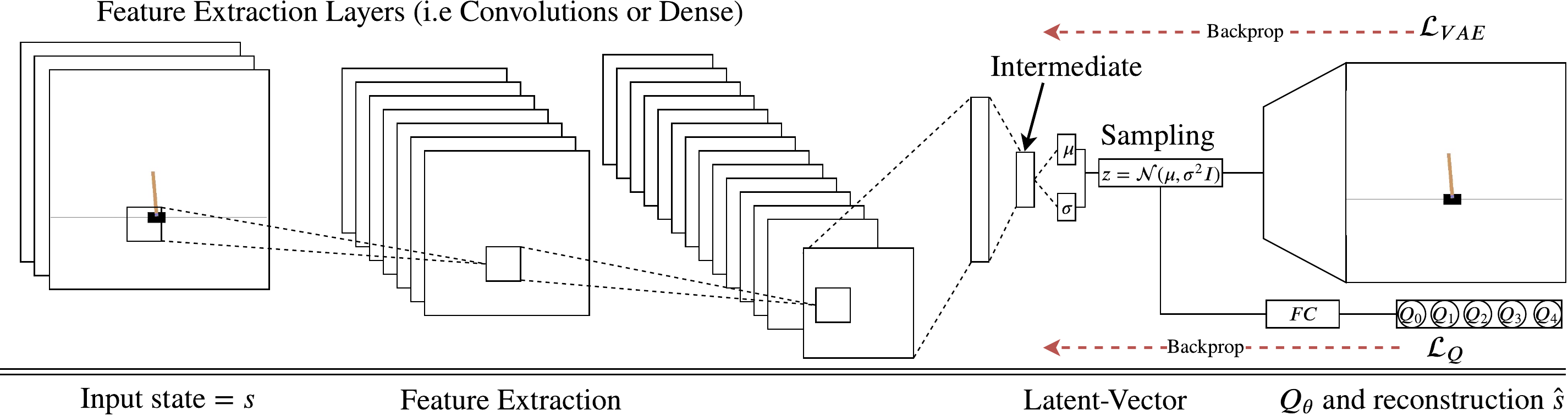}
    \caption{The deep variational Q-Networks architecture.}
    \label{fig:dvqn}
\end{figure*}

Figure \ref{fig:dvqn} illustrates the architecture of the algorithm. The architecture follows general trends in similar RL literature but has notable contributions. First, features are extracted from the state-input, typically by using convolutions for raw images and fully-connected for vectorized input. The extracted features are forwarded to a fully connected intermediate layer of a user-specified size commonly between 50 to 512 neurons. The intermediate layer splits into two streams that represent the variance \(\mu\) and standard deviation \(\sigma\) and is used to sample the latent-vector using a Gaussian distribution through the re-parameterization. The latent-vector is forwarded to the decoder for state reconstruction and the Q-Learning stream for action-value (Q-value) optimization. The decoder and Q-Learning streams have the following loss functions: 

\begin{equation}
 \mathcal{L}_{VAE} = MSE(s, \hat{s}) + D_{KL}[q_\psi(z|s) \| p_\theta(z|s)]
\end{equation}
\begin{equation}
\mathcal{L}_{DQN} =( r  + \gamma Q(s', \arg \max_{a'}Q(s', a'; \theta_i) ;\theta_i)- Q(s, a; \theta_i))^2
\end{equation}
\begin{equation}
\mathcal{L}_{DVQN} = c_1\mathbb{E}_{\sim q_\psi(z|s) }[ \mathcal{L}_{VAE}] + c_2\mathbb{E}_{s, a, s', D  \sim r}[\mathcal{L}_{DQN}].
\label{eq:full_loss}
\end{equation}

The global loss function \(\mathcal{L}_{DVQN}\) is composed of two local objectives: \(\mathcal{L}_{DQN} \) and \(\mathcal{L}_{VAE} \). In the VAE loss, the first term is the mean squared error between the input \(s\) and its reconstruction \(\hat{s}\). The second term is regularization using KL-distance to minimize the distance between the latent distribution and a Gaussian distribution. The DQN loss is a traditional deep Q-Learning update, as described in \cite{Mnih2015}.

\begin{algorithm}[H]
    \caption{DVQN: Minimal Implementation}
    \label{alg:dvqn} 
    \begin{algorithmic}[1]
        \State Initialise  $\Omega$
        \State Initialise DVQN model $\pi$
        \State Initialise replay-buffer \(D_\pi\)
        
        \For {N episodes}
        \State \(D_{\pi} \leftarrow\) Collect samples from \(\Omega\) under the untrained policy \(\pi\) via the generative policy sampling.
        \State Train model \(\pi\) on a mini-batch from \(D_{\pi}\) with objective from Equation \ref{eq:full_loss}
        \EndFor
    \end{algorithmic}
\end{algorithm}

Algorithm \ref{alg:dvqn} shows a general overview of the algorithm. First, the environment is initialized. Second, the DVQN model from Figure \ref{fig:dvqn} is initialized with the desired hyperparameters, and third, the replay-buffer is created. For a specified number of episodes, the algorithm samples actions from the generative policy for exploration and stores these as MDP tuples in the experience replay. After each episode, the algorithm samples mini-batches from the experience replay and performs parameter updates using stochastic gradient descent. The (loss function) optimization objective is described in equation \ref{eq:full_loss}. The process repeats until the algorithm converges.

\section{Experiments and Results}
\label{sec:results}
In this section, we conduct experiments against four traditional environments to demonstrate the effectiveness of the DVQN algorithm. We show that the algorithm can organize the latent-space by state similarity while maintaining comparable performance to model-free deep Q-learning algorithms. 

\subsection{Experiment test-bed}
\begin{figure}
    \centering
    \includegraphics*[width=0.6\linewidth]{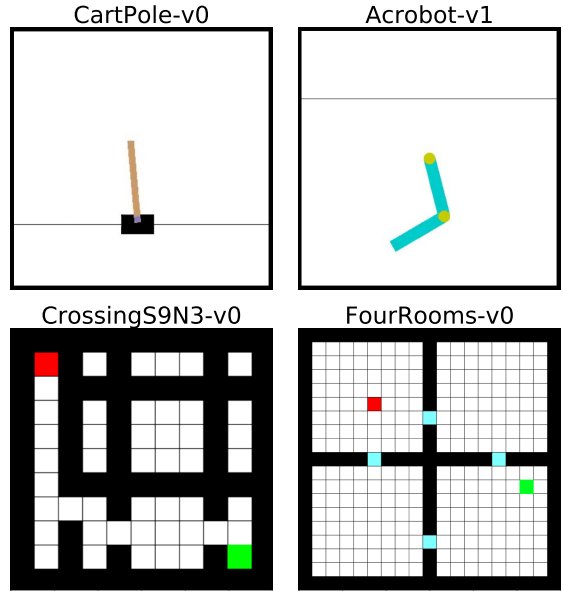}
    \caption{The experiment test-bed contains the following environments; CartPole-v0, Acrobot-v1, CrossingS9N3-v0, and FourRooms-v0}
    \label{fig:environments}
\end{figure}

We evaluate the DVQN in the following environments; CartPole-v0, Acrobot-v1,  CrossingS9N3-v0, and FourRooms-v0, shown in Figure \ref{fig:environments}. These environments are trivial to solve using model-free reinforcement learning and hence, excellent for visualizing the learned latent-space. The FourRooms-v0 environment is especially suited for option-based RL and can solve the problem in a fraction of time steps compared to model-free RL. Although the DVQN algorithm \textbf{does not quantify options for analysis} (see Section \ref{sec:conclusion}), the primary goal is to organize the latent-space so that it is possible to extract meaningful and interpretable options automatically. The aim is to have \textbf{comparable performance} to  vanilla deep Q-Network variants found in the literature \cite{Mnih2015,VanHasselt2015,Hessel2017}. DVQN benchmarks against vanilla DQN, Double DQN, and Rainbow.

FourRooms-v0 and CrossingS9N3-v0 are a grid-world environment where the objective is to reach the terminal-state cell (In the lower right of the image in both environments). In FourRooms-v0, the agent has to enter several doors and only complete a part of the goal for each door it enters. FourRooms-v0 is an ideal environment for option-based reinforcement-learning because each door is considered a sub-goal. While the environment is solvable by many deep reinforcement learning algorithms, option-based RL is more efficient. The agent receives small negative rewards for moving and positive rewards for entering the goal-state (global) or the doors (local). The Crossing is a simpler environment where the agent has to learn the shortest path to the goal state. In both grid-environments, the agent can move in any direction, one cell per time step.

To further show that the algorithm works in simple control tasks, we perform experiments in CartPole-v0 and Acrobot-v1. The objective in CartPole-v0  is to balance a pole on a cart. Each step generates a positive reward signal while receiving negative rewards if the pole falls below an angle threshold. The agent can control the direction of the cart at every time step. The Acrobot-v1 has a similar aim to control the arm to hit the ceiling in a minimal number of time steps. The agent receives negative rewards until it reaches the ceiling. The CartPole-v0 and Acrobot-v1 environments origins from \cite{Brockman2016a} while CrossingS9N3-v0 and the FourRooms-v0 origins from \cite{gym_minigrid}.\footnote{A community-based scoreboard can be found at \url{https://github.com/openai/gym/wiki/Leaderboard}.}

\subsection{Hyperparameters}

\begin{table*}[t]
    \centering
    \begin{tabular}{|l|c|l|l|c|}
        \hline
\textbf{Algorithm}     & \multicolumn{1}{l|}{\textbf{DQN}} & \textbf{DDQN} & \textbf{Rainbow} & \multicolumn{1}{l|}{\cellcolor[HTML]{EFEFEF}\textbf{DVQN (ours)}} \\ \hline
\textbf{Optimiser}     & \multicolumn{3}{c|}{Adam}                                            & \cellcolor[HTML]{EFEFEF}RMSProp                                   \\ \hline
\textbf{Learning Rate} & \multicolumn{3}{c|}{0.003}                                           & \cellcolor[HTML]{EFEFEF}0.000025                                  \\ \hline
\textbf{Activation}    & \multicolumn{3}{c|}{ReLU}                                            & \cellcolor[HTML]{EFEFEF}ELU                                       \\ \hline
\textbf{Batch Size}    & \multicolumn{3}{c|}{32}                                              & \cellcolor[HTML]{EFEFEF}128                                       \\ \hline
\textbf{Replay Memory} & \multicolumn{4}{c|}{1m}                                                                                                                  \\ \hline
\textbf{Epsilon Start} & \multicolumn{3}{c|}{1.0}                                             & \cellcolor[HTML]{EFEFEF}N/A                                       \\ \hline
\textbf{Epsilon End}   & \multicolumn{3}{c|}{0.01}                                            & \cellcolor[HTML]{EFEFEF}N/A                                       \\ \hline
\textbf{Epsilon Decay} & \multicolumn{3}{c|}{0.001 (Linear)}                                  & \cellcolor[HTML]{EFEFEF}N/A                                       \\ \hline
\textbf{Gamma}         & \multicolumn{4}{c|}{0.95}                                                                                                                \\ \hline
\textbf{Q-Loss}        & \multicolumn{3}{c|}{Huber}                                                       & \cellcolor[HTML]{EFEFEF}MSE                                       \\ \hline
    \end{tabular}
    \caption{Algorithm and hyperparameters used in the experiments. For the Rainbow algorithm, we used the same hyperparameters described in \cite{Hessel2017}. The DDQN had target weight updates every 32k frames.}
    \label{tab:hyperparamters}
\end{table*}

During the experiments, we found DVQN to be challenging to tune. Initially, the algorithm used ReLU as activation but was discarded due to vanishing gradients resulting in divergence for both policy and reconstruction objectives. By using ELU, we found the algorithm to be significantly more stable during the experiments, and it additionally did not diverge if training continued after convergence. We will explore the underlying cause of our future work. Table \ref{tab:hyperparamters} shows the hyperparameters used in our experiments where most of the parameters are adopted from prior work. Recognize that the DVQN algorithm does not use \(\epsilon\)-greedy methods for exploration. The reason for this is that random sampling is done during training in the variational autoencoder part of the architecture. In general, the algorithm tuning works well across all of the tested domains, and better results can likely be achieved with extended hyperparameter searches. For DVQN, we consider such tuning for our continued work on DVQN using options, see Section \ref{sec:conclusion}.

\subsection{Latent-Space evaluation}

\begin{figure}
    \centering
    \includegraphics[width=\linewidth]{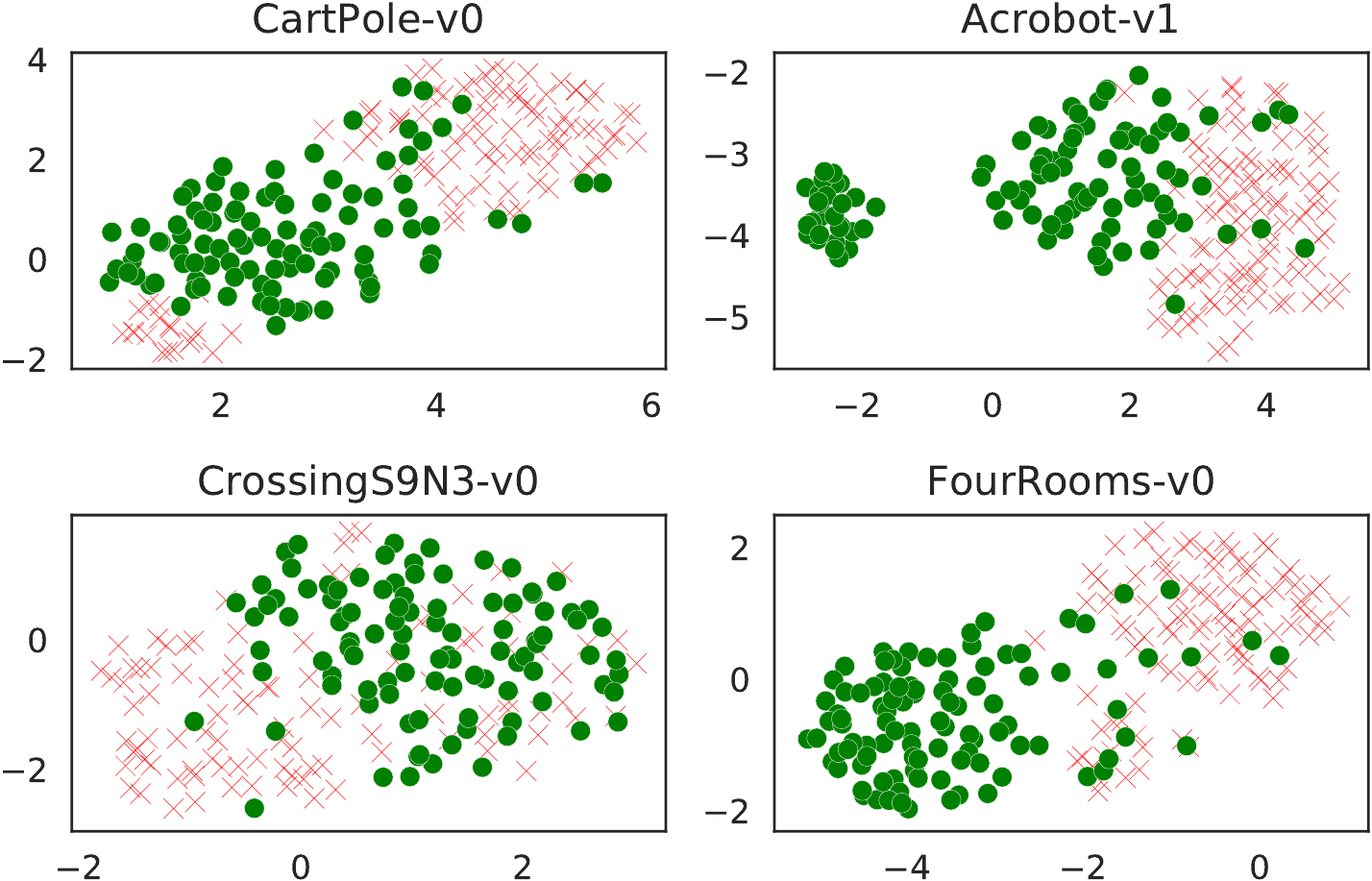}
    \caption{The learned latent space for all of the tested environments. DVQN successfully trivialise the selection of options as seen in the well-separated state clusters. The circular points illustrate states with positive reward while cross illustrates negative rewards. }
    \label{fig:res_embedding}
\end{figure}

\begin{figure}
    \centering
    \includegraphics[width=0.8\linewidth]{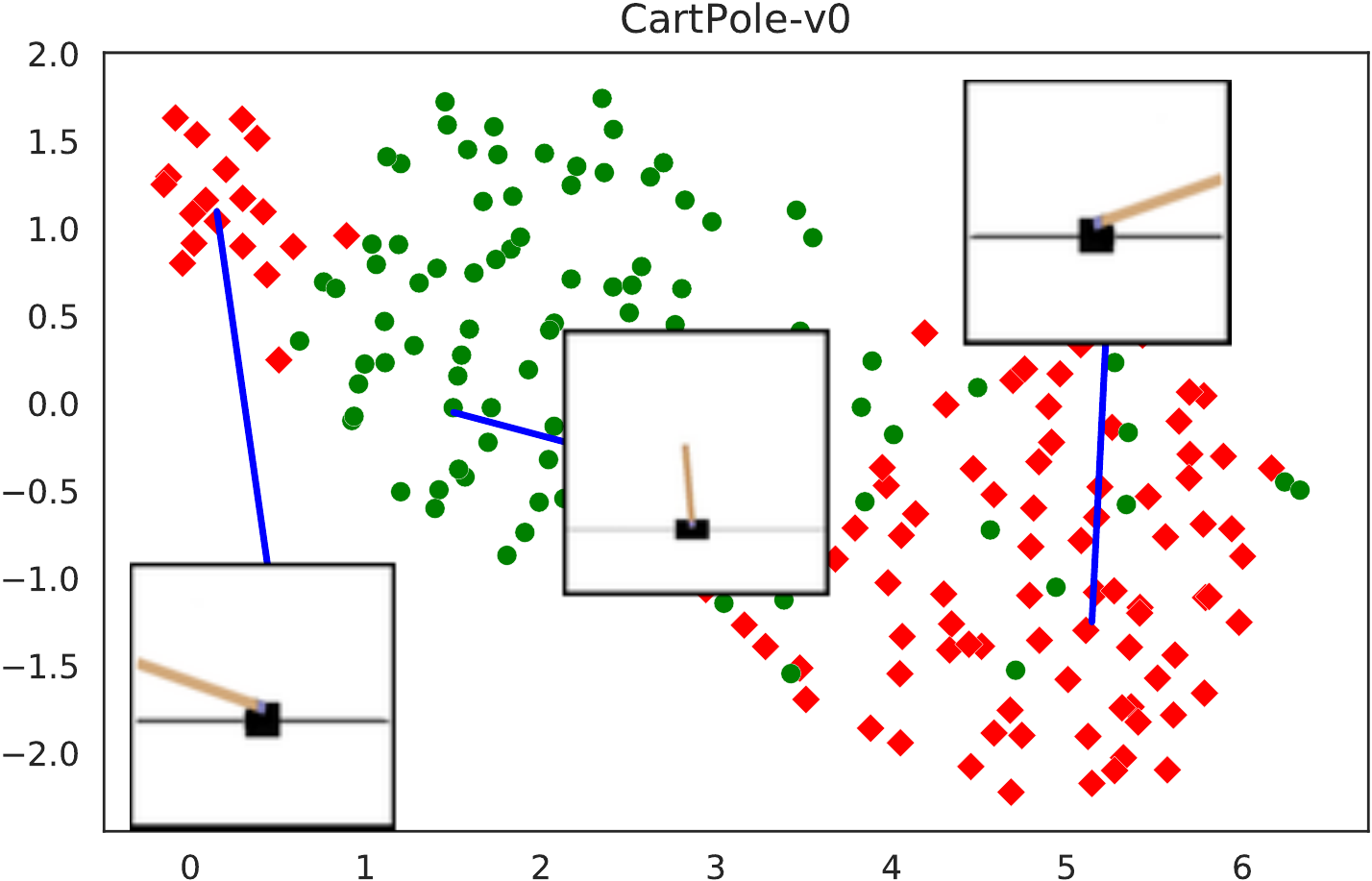}
    \caption{The relationship between states and the latent-space for the CartPole-v0 environment. DVQN can separate each angle, left, middle and right into separable clusters, which are especially useful in option-based reinforcement learning. Additionally, the visualization of the latent space that the Q-head uses to sample actions is trivial to interpret.}
    \label{fig:res_embedding_detail}
\end{figure}

An attractive property of our model is that the latent-space is a Gaussian distribution. As seen in Figure \ref{fig:res_embedding}, the DVQN algorithm can produce clustered latent-spaces for all tested environments. For example, in the CartPole-v0 environment, there are three clusters where two of them represent possible terminal states and one that represents states that give a reward. To fully utilize the capabilities of DVQN, the latent-space can be used to generate options for each cluster to promote different behavior for every region of the state-space.

Figure \ref{fig:res_embedding_detail} illustrates the visualization of the latent-space representation in CartPole-v0. We find that each cluster represents a specific position and angle of the pole. The latent-space interpolates between these state variations, which explains its shape. Although the clusters are not perfect, it is trivial to construct separate classification for each cluster with high precision, and this way automatically construct initiation and termination signals for options.

\subsection{Performance evaluation}

\begin{figure*}
    \centering
    \includegraphics*[width=\linewidth]{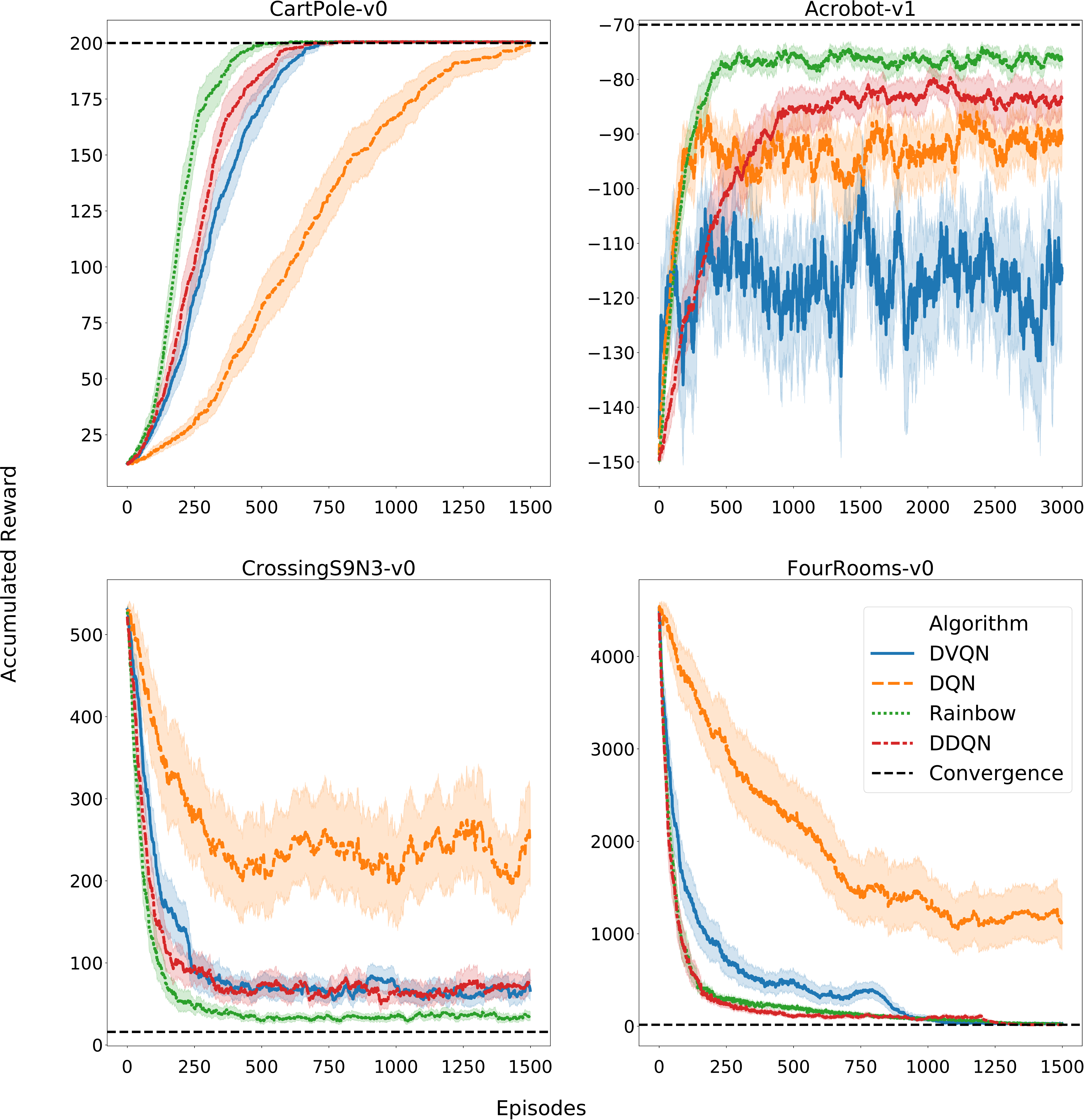}
    \caption{The accumulative sum of rewards of the DVQN compared to other Q-Learning based methods in the experimental environments. Our algorithm performs better than DQN from \cite{Mnih2015}, and shows comparable results to DDQN from \cite{VanHasselt2015a} and Rainbow  from \cite{Hessel2017}. We define an episode threshold for each of the environments (x-axis) and accumulate the agent rewards as the performance metric for CartPole-v0 and Acrobot-v1. The scoring metric in CrossingS9N3-v0 and the FourRooms-v0 is based on how many steps the agent used to reach the goal state.}
    \label{fig:performance}
\end{figure*}

Figure \ref{fig:performance} illustrates a comparison of performance between state-of-the-art Q-Learning algorithms and the proposed DVQN algorithm. The performance measurement is the mean of 100 trials over 1500 episodes for CartPole-v0, \\CrossingS9N3-v0, FourRooms-v0, and 3000 episodes for Acrobot-v1. The performance is measured in accumulated rewards and is therefore negative for environments where each time step yields a negative reward.

The DVQN algorithm performs better than DQN and shows comparable performance to DDQN and Rainbow. DVQN is not able to find a good policy in the Acrobot-v1 environment but successfully learns a good visual representation of the latent space. In general, the DVQN algorithm is significantly harder to train because it requires the algorithm to find a good policy within a Gaussian distribution. We found this to work well in most cases, but it required fine-tuning of hyperparameters. The algorithm is also slower to converge, but we were able to improve training stability by increasing the batch-size and decreasing the learning-rate.

\section{Conclusion and Future Work}
\label{sec:conclusion}
This paper introduces the deep variational Q-network (DVQN), a novel algorithm for learning policies from a generative latent-space distribution. The learned latent-space is particularly useful for clustering states that are close to each other for \textbf{discovering options automatically}. In the tested environments, the DVQN algorithm can achieve \textbf{comparable performance} to traditional deep Q-networks. DVQN does not provide the same training stability and is significantly harder to fine-tune than traditional deep Q-learning algorithms. For instance, network capacity is increased. As a result of this, the algorithm takes longer to train, and during the experiments, only the RMSprop optimizer \cite{Nair2010} with a small step size was able to provide convergence. Additionally, the exponential linear units from \cite{Clevert2015} had a positive effect on stability. On the positive side, the DVQN contributes a \textbf{novel approach for options discovery} in hierarchical reinforcement learning algorithms.

The combination of VAE and reinforcement learning algorithms has interesting properties. Under \textbf{optimal conditions}, the latent-space should, in most cases follow a true Gaussian distribution where policy evaluations always provide optimal state-action values, since this is the built-in properties of the latent space in any VAE. The difference between traditional deep Q-networks and DVQN primarily lies in the elimination of a sparse and unstructured latent-space. In deep Q-Networks, optimization does not provide a latent-space structure that reflects a short distance between states but rather a distance between Q-values \cite{Mnih2015}. By using KL-regularization from VAE, low state-to-state is encouraged. Another benefit of VAE is that we sample from a Gaussian distribution to learn \(\mu\) and \(\sigma\), which is especially satisfying for algorithms with off-policy sampling and therefore eliminates the need for (\(\epsilon\)-greedy) random sampling.

In the continued work, we wish to do a thorough analysis of the algorithm to justify its behavior and properties better. A better understanding of the Gaussian distributed latent-space is particularly appealing because it would enable better labeling schemes for clustering, or perhaps fully automated labeling. Finally, we plan to extend the algorithm from model-free behavior to hierarchical RL with options. The work of this contribution shows that it is feasible to produce organized latent-spaces that could provide meaningful options, and the hope is that this will result in state-of-the-art performance in a variety of tasks in RL.

\bibliographystyle{splncs04}

\end{document}